\documentclass[letterpaper, 10 pt, conference]{ieeeconf}  % Comment this line out if you need a4paper

\IEEEoverridecommandlockouts    % This command is only needed if 
\usepackage{balance}

\usepackage{graphics} % for graphics files
\usepackage{graphicx} % Per figure
\usepackage{wrapfig} % Per figure con caption

\usepackage{mathptmx} % assumes new font selection scheme installed
\usepackage{times} % assumes new font selection scheme installed
\usepackage{amsmath} % assumes amsmath package installed
\usepackage{amssymb}  % assumes amsmath package install
\usepackage{amsfonts} % Roba matematica

\usepackage{verbatim} % comments
\usepackage{siunitx} % Unit of measure
\usepackage{float} % float figures
\usepackage{subcaption} % sub figures with captions
\usepackage[hidelinks]{hyperref}
\usepackage{url} % Include gli url
\usepackage{booktabs}
\usepackage[utf8]{inputenc} % Encoding
\usepackage[english]{babel} % Lingua

\usepackage{multirow} % Tabelle non omogenee
\usepackage{tabularx} % Tabelle a dimesione fissa e custom
\newcolumntype{Y}{>{\centering\arraybackslash}X} % Colonne centrate lunghezza dinamica
\usepackage{color} %colori

\usepackage{mathtools} % for prescript
\usepackage{textcomp} %for tilde

\title{\LARGE \bf
Vision-based Control of a Quadrotor in User Proximity:\\Mediated vs End-to-End Learning Approaches
}

\author{Dario Mantegazza, J\'er\^ome Guzzi, Luca M. Gambardella and Alessandro Giusti % <-this % stops a space
\thanks{All authors are with the Dalle Molle Institute for Artificial Intelligence (IDSIA), USI-SUPSI, Lugano, Switzerland. This work is supported by the Swiss National Science Foundation (SNSF) through the NCCR Robotics.}% <-this % stops a space
}

\usepackage{lipsum}

\begin{document}

\maketitle
\thispagestyle{empty}
\pagestyle{empty}

%%%%%%%%%%%%%%%%%%%%%%%%%%%%%%%%%%%%%%%%%%%%%%%%%%%%%%%%%%%%%%%%%%%%%%%%%%%%%%%%
\begin{abstract}
We consider the task of controlling a quadrotor to hover in front of a freely moving user, using input data from an onboard camera.  On this specific task we compare two widespread learning paradigms: a mediated approach, which learns a high-level state from the input and then uses it for deriving control signals; and an end-to-end approach, which skips high-level state estimation altogether.  We show that despite their fundamental difference, both approaches yield equivalent performance on this task. We finally qualitatively analyze the behavior of a quadrotor implementing such approaches.
\end{abstract}

\section*{Videos, Datasets, and Code}
Videos, data, and code to reproduce our results are available at: \url{https://github.com/idsia-robotics/proximity-quadrotor-learning}.

%%%%%%%%%%%%%%%%%%%%%%%%%%%%%%%%%%%%%%%%%%%%%%%%%%%%%%%%%%%%%%%%%%%%%%%%%%%%%%%%
\section{Introduction}
\label{sec:introduction}
Robot control systems are traditionally structured in two distinct modules: perception and control.  Perception processes the robot’s input in order to derive a high-level state, which represents meaningful and relevant information for the task the robot needs to solve. State information is then used by a controller, which determines the low-level control signals to be provided to the hardware.

For mobile robots operating in real-world unstructured environments, perception is often challenging. This is especially true when this step involves the interpretation of complex, high-dimensional data such as images.  Many recent successful systems deal with this problem by adopting supervised machine learning (ML) techniques which operate on sensing data as input.  %In particular, deep learning techniques have repeatedly demonstrated excellent results on raw sensor data.

%We consider the design of such a solution for a specific task, which is challenging but conceptually simple: controlling a quadrotor to fly in order to stay 1.5~m in front of the face of a freely moving user, using odometry and raw frames from a forward-looking camera as input.  In this context, the high-level state is given by the pose of the user's head relative to the drone.  The control signals are desired pitch, roll, yaw velocity and vertical speed.

When designing such a system, we face the choice between at least two approaches.
\begin{itemize}
\item A mediated approach, in which one trains a supervised ML model to predict the high-level state given the inputs; then, control signals are derived from the state using a designed controller (or another learned model).
\item An end-to-end approach, in which a supervised ML model is trained to directly predict the control signals from the sensing data, without passing through an intermediate high-level representation of the state.
\end{itemize}

%\begin{figure}
%    \centering
%    \missingfigure[figwidth=\columnwidth]{intro}
%    \caption{Control architectures}
%    \label{fig:intro}
%\end{figure}

Which architecture is preferable? In some situations, one might not have a choice: for example, if one can not (or does not want to) collect ground truth information about the high-level state, training a perception model for a mediated approach is not possible (and this fact motivates many end-to-end systems). However, sometimes datasets annotated with such a ground truth can be acquired, potentially at a cost.  Is it worth it?

There are several advantages in a mediated approach: 1) in many mobile robotics applications, hand-designing a controller, given a high-level state representation, is feasible and gives the designer explicit control on the resulting robot behavior; 2) a mediated approach is more transparent and may be easier to inspect and debug: given an unexpected robot behavior, the designer can inspect the high-level state to determine whether the problem is in the perception or controller.

In contrast, end-to-end approaches are appealing because they are conceptually very simple and can potentially be more computationally efficient, especially if the high-level state representation is complex and high-dimensional.  Moreover, as mentioned above, end-to-end approaches don’t depend on high-level state's ground truth for training.

The considerations above disregard one key issue, i.e., the difficulty of learning models for the two approaches.  Is learning a perception model (which outputs high-level state) easier than learning an end-to-end model (which directly outputs control signals)?  Does one of the two models require a larger amount of training data to reach the same performance?

% [FIXME Motivate the reasons why we may think one way or the other.  Needs an example though.]

%This paper investigates this question in the context of one specific task, which is challenging but conceptually simple: controlling a quadrotor to fly in order to stay 1.5~m in front of the face of a freely moving user, using odometry and raw frames from a forward-looking camera as input. The high-level state is given by the pose of the user's head relative to the drone.  The control signals are desired pitch, roll, yaw velocity and vertical speed.

This is a key issue to consider when designing ML-based robot controllers: we investigate this question for one specific task. After reviewing related literature (Section~\ref{sec:rw}), we model and formalize the end-to-end architecture and two variants of the mediated architecture (Section~\ref{sec:model:general}); we instantiate these three architectures for one specific task, that we consider in the remainder of the paper, i.e., controlling a quadrotor in order to hover in front of a person who is freely moving (Section~\ref{sec:model}); such behavior could be implemented by a quadrotor tasked to monitor a person or expecting commands from them. The \textbf{main contribution} of the paper is a set of experiments showing that, for this specific task, training models for the three architectures has the same difficulty (the setup and results for these experiments are described in Sections ~\ref{sec:setup} and \ref{sec:results} respectively).  The \textbf{main limitation} is that we limit our study to a single reactive control task; nonetheless, it is a challenging, real-world task which has several characteristics in common with other important subproblems in mobile robotics. %[FIXME a bit abrupt]. 
A \textbf{secondary contribution} of this paper is the design, implementation and validation of the drone control system described above, which constitutes a useful component in applications involving proximal interaction of humans and quadrotors; collected datasets, source code and training models are available for download.

\subsection{Related Work}
\label{sec:rw}
End-to-end learning approaches map raw sensor data (lidar, camera images, ego-motion sensors) to control actions: either as direct low-level control outputs~\cite{LeCun2005}, or as target points (e.g., desired velocity) for optimal low-level controllers~\cite{Kaufmann2018}.  

Supervised end-to-end approaches have been successfully applied to a variety of challenging control scenarios: off-road obstacle avoidance~\cite{LeCun2005, Bajracharya2008}; autonomous driving~\cite{Chen2017,Xu2017,Chi2017,Heylen2018,bojarski2016end}; vision-based manipulation~\cite{Levine2016}; quadrotor control in forested environments~\cite{Ross2013,Giusti2016,Smolyanskiy2017}, cities~\cite{Loquercio2018,Kumaar2018}, and cluttered environments~\cite{Kaufmann2018}.

These can be considered instances of imitation learning (also known as learning-from-demonstration): in fact, one trains a policy to solve a sequential task using only demonstrations by an expert (represented as state/action pairs), as training data~\cite{attia2018global}. In end-to-end approaches, the state coincides with the robot sensing; this yields reactive controllers unless recurrent models are used to capture temporal dynamics~\cite{Chi2017}.

The ground truth used for learning may have various origins: skilled drivers or operators~\cite{Ross2013,Kumaar2018,LeCun2005,Bajracharya2008}; people walking~\cite{Giusti2016,Smolyanskiy2017} or driving a vehicle that is not the target robot~\cite{Loquercio2018}; random controllers that sometimes lead to collisions, which the model learns to avoid~\cite{Gandhi2017}; hand-designed controllers~\cite{Kaufmann2018}; %applied to the exact state, which may be returned by an external motion tracker, or even a
controllers learned through reinforcement learning~\cite{Levine2016}; or the future position of vehicles in a large driving dataset~\cite{Xu2017}.
Most of these approaches are stateless and reactive, but applications of Deep Recurrent Neural Networks allows to capture temporal dynamics in end-to-end approaches~\cite{Chi2017}.

Compared to mediated approaches, end-to-end learning has been found in some cases to be slower to converge~\cite{Glasmachers2017,Shalev2017} and to require more training samples~\cite{Shalev2016}. Direct perception methods~\cite{Chen2015} are a further alternative for learning higher-level representations of the environment than mediated methods.

The task we consider in this paper is aimed to proximity human-robot interaction~\cite{Ng2011,Naseer2013}: a drone should be able to fly at an appropriate distance in front of people, following them~\cite{Jahidul2018} while waiting for possible command gestures~\cite{Peshkova2017}. The use of visual markers~\cite{Vasconcelos2016} simplifies this task but, in general, more sophisticated techniques, like Tracking-Learning-Detection~\cite{Kalal2012}, are needed~\cite{Bartal2015}. For instance, several deep learning approaches have been proposed to estimate the 6D pose of a person's head~\cite{Patacchiola2017} or body~\cite{Bogo2016} from monocular cameras (as well of general objects~\cite{Xiang2017}).  Adopting such perception modules would be a reasonable alternative to solve our task; however, in this paper the task acts as a model of a larger class of tasks.

% As a case study for the comparison between mediated and end-to-end approaches, we present the first end-to-end solution to let a quadrotors follow people while staying in front of them.

\section{Task and Model}
\label{sec:model}

\newcommand{\xall}{\mathbf{x}}
\newcommand{\xim}{\mathbf{x}_\text{im}}
\newcommand{\xodom}{\mathbf{x}_\text{odom}}
\newcommand{\uall}{\mathbf{u}}
\newcommand{\uax}{u_{\text{a}x}}
\newcommand{\uay}{u_{\text{a}y}}
\newcommand{\uvz}{u_{\text{v}z}}
\newcommand{\uoz}{u_{\omega z}}
\newcommand{\sall}{\mathbf{s}}
\newcommand{\spose}{\mathbf{s}_\text{pose}}
\newcommand{\sodom}{\mathbf{s}_\text{odom}}
\newcommand{\sx}{s_x}
\newcommand{\sy}{s_y}
\newcommand{\sz}{s_z}
\newcommand{\st}{s_\theta}
\newcommand{\controller}{C}
\newcommand{\controlleracq}{C_\text{acq}}

%We first describe the specific task that we are considering.  Then, we outline the three general solution approaches.  Finally, we discuss the relevance of these approaches to other tasks.

%\subsection{Task}
%\label{sec:model:task}
We consider one specific task: controlling a quadrotor to stay at a fixed distance ($\Delta =\SI{1.5}{m}$) and at eye-level height in front of a user who is free to move in an environment.

% Inputs and outputs
The available inputs are the video feed from a forward-pointing camera $\xim$ and the current linear velocity of the quadrotor $\xodom = (v_x, v_y)$, obtained through ego-motion sensors;
%fed with data acquired by a bottom-looking camera
the controller outputs $\uall$ are composed by:
\begin{itemize}
    \item the desired pitch and roll of the drone, which map to the acceleration along the drone's $x$ and $y$ axes ($\uax$, $\uay$ respectively);
    \item the desired velocity along the $z$ axis ($\uvz$);
    \item the desired angular velocity around the $z$ axis ($\uoz$).
\end{itemize}

% State
The state $\sall$ for the given task can be compactly represented using two pieces of information $\sall = (\spose, \sodom)$.
\begin{itemize}
\item A 3D transformation $\spose$, representing the pose (location and orientation) of the user's head with respect to the quadrotor; more specifically, only the heading component of the head's orientation is relevant to the task (in fact, tilting the head or looking up or down should not affect the drone behavior).  Therefore, we represent with angle $\st$, $-\pi \leq \st \leq \pi$, the relative angle between the user's and drone's orientation, such that $\st = 0$ if the user's face points along the negative $x$ axis of the drone (i.e., the user is in front of the drone and faces towards it).  Therefore, $\spose = (\sx, \sy, \sz, \st)$.
\item In addition, the state also contains the quadrotor's current measured velocity, which is available directly as an input $\sodom = \xodom$.
\end{itemize}

In this specific task, we can also design a controller $\controller$ that produces control signals given the state: $\uall = f_{\controller}(\sall)$, as detailed in Section~\ref{sec:controller}. Note that for other tasks (e.g. grasping), designing a controller might be very complex.

\subsection{General Approaches}

\label{sec:model:general}
\begin{figure}[ht]
    \centering
    \includegraphics[width=\columnwidth]{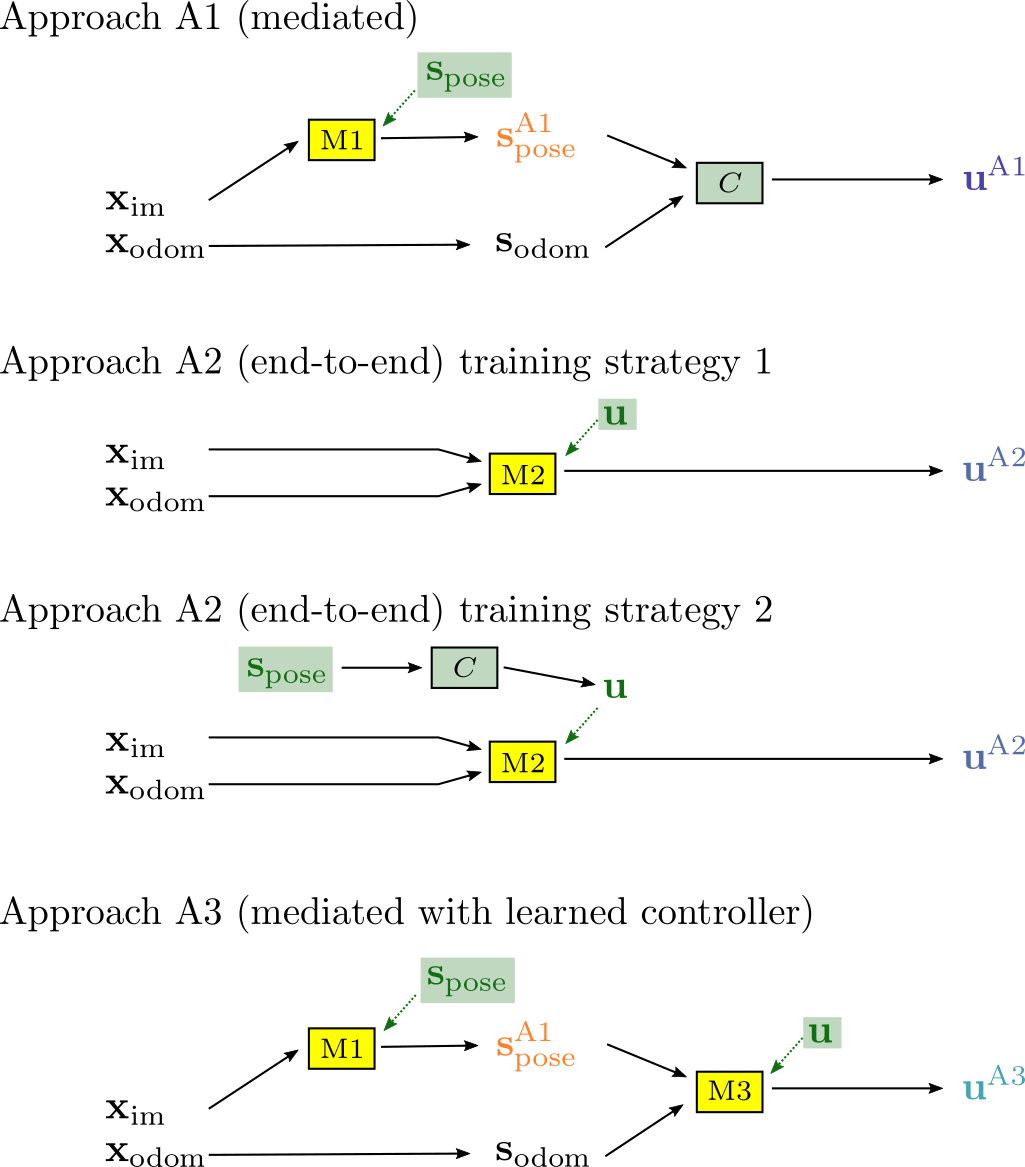}
    \caption{Yellow background: learned models. Ground truth data used for training them is depicted with a dotted arrow. Requirements (in terms of ground truth or of a designed controller) are drawn with a green background.}
    \label{fig:approaches}
\end{figure}

We now describe the three approaches that we will compare in the following sections (see Figure~\ref{fig:approaches}).  Even though for clarity we use the task-specific notation introduced above, these approaches can be easily generalized to other control tasks; then, $\xim$ corresponds to the subset of the inputs that we want to learn how to interpret in order to obtain high-level state information $\spose$.  $\xodom$ represents inputs that we can (directly or through some processing step) turn into meaningful high-level state information $\sodom$.  For example, in an autonomous driving scenario: $\xim$ could correspond to lidar data; $\spose$ to the relative position of other cars on the road and their velocities; $\sodom$ to the car odometry and to the car's position on the lane (obtained through some existing modules using inputs from $\xodom$); $\uall$ to steering and acceleration/brake controls. 
% Different approaches have different requirements (summarized in Table~\ref{tab:req}).

In \textbf{approach A1 (mediated)}, we learn a model M1 mapping $\xim$ to $\spose$. Let $f_\text{M1}$ denote the function implemented by M1; $\spose^\text{A1} = f_\text{M1}(\xim)$ denotes the estimate of $\spose$.  $\spose^\text{A1}$ is then joined to the odometry (available directly in the input) to form $\sall^\text{A1} = (\spose^\text{A1}, \sodom)$. The control signals are then computed as $\uall^\text{A1} = f_{\controller}(\sall^\text{A1})$.  In order to train M1, we require ground truth information on $\spose$.  Moreover, this approach requires the availability of a controller $\controller$.

In \textbf{approach A2 (end-to-end)}, we directly learn a function $f_\text{M2}$ to predict $\uall$ from $\xall = (\xim, \xodom)$.  We denote the result as
$\uall^\text{A2} = f_\text{M2}(\xall)$.  Training M2 requires that a ground truth for $\uall$ is available.  Such a ground truth can be obtained in two ways.  \emph{Learning strategy 1}: acquiring the ground truth for $\uall$ directly (e.g., recording a skilled human pilot); \emph{Learning strategy 2}: obtaining the ground truth for $\uall$ through a designed controller $\controller$, which is given ground truth $\spose$ information.

In \textbf{approach A3 (mediated with learned controller)} we learn a model M1 just like in approach A1.  This model obtains a state estimate $\sall^\text{A3} = \sall^\text{A1}$.  However, instead of using an hand-designed controller $\controller$ to produce the control signals from the estimated state, we learn a mapping $f_\text{M3}$ from $\sall$ to $\uall$.  The control signals are then computed as $\uall^\text{A3} = f_\text{M3}(\sall^\text{A3})$.  In order to train M1, we require ground truth information on $\spose$. Moreover, in order to train M3 we need a ground truth on $\uall$, e.g. from a human controller.  Differently from A1, this approach does not require one to design the controller $\controller$.

%\subsection{Discussion on approaches}
%Depending on the scenario, only some of the approaches above may be viable.  In particular, the three approaches have different requirements, summarized in Table~\ref{tab:req}

\begin{comment}
\begin{table}[]
\centering
\caption{Approach requirements}
\label{tab:req}
\begin{tabular}{@{}llll@{}}
\toprule
Approach     & State labels & Controller $\controller$ & Output labels              \\ \midrule
A1           & Required     & Required            & Not required \\
A2 strategy 1 & Not required & Not required        & Required               \\
A2 strategy 2 & Required     & Required            & Not required \\ 
A3           & Required     & Not required        & Required \\ \bottomrule
\end{tabular}
\end{table}
\end{comment}

\begin{figure*}[t]
    \centering \vspace{2mm}
    \begin{subfigure}[b]{0.32\textwidth}
        \includegraphics[width=\textwidth]{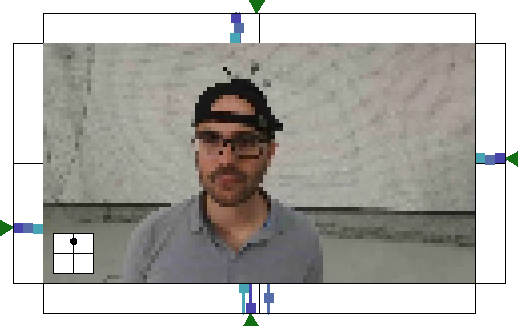}
        \label{fig:0_test}
    \end{subfigure}
    \begin{subfigure}[b]{0.32\textwidth}
        \includegraphics[width=\textwidth]{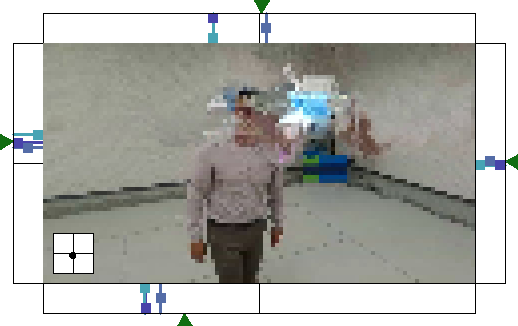}
        \label{fig:3_test}
    \end{subfigure}
    \begin{subfigure}[b]{0.32\textwidth}
        \includegraphics[width=\textwidth]{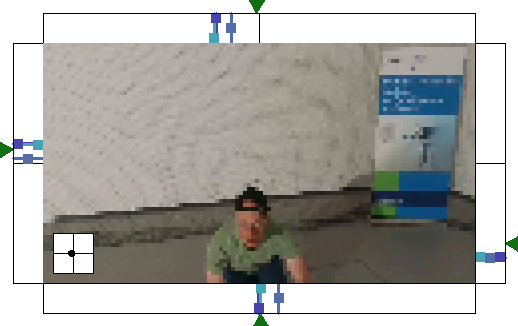}
        \label{fig:9_test}
    \end{subfigure}
    \begin{subfigure}[b]{0.32\textwidth}
        \includegraphics[width=\textwidth]{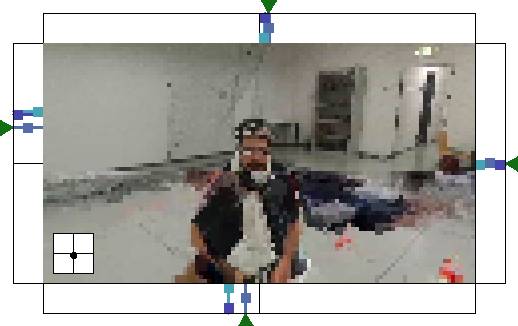}
        \label{fig:6_test}
    \end{subfigure}
    \begin{subfigure}[b]{0.32\textwidth}
        \includegraphics[width=\textwidth]{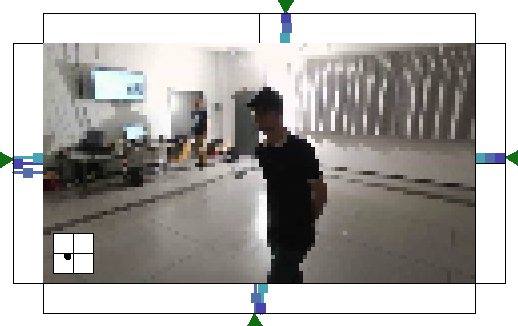}
        \label{fig:11_test}
    \end{subfigure}
    \begin{subfigure}[b]{0.32\textwidth}
        \includegraphics[width=\textwidth]{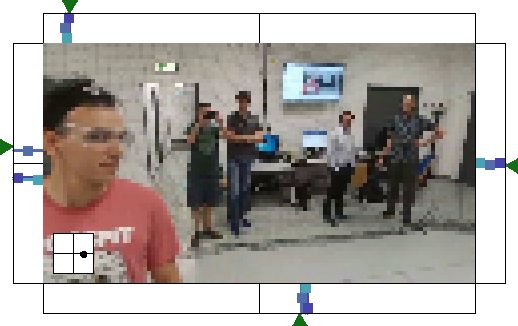}
        \label{fig:12_test}
    \end{subfigure}

    \caption{Instances from the dataset. Each shows ground truth (green triangle) and outputs (blue rectangles) of A1, A2, A3 for each control variable: $\uax$ (left), $\uay$ (bottom), $\uvz$ (right), $\uoz$ (top).  More details and videos in supplementary material. Note that some images in the dataset are affected by acquisition, compression or transmission artifacts.%For each we display: a) The video frame $\xim$, in the center.  b) The measured velocity $\xodom$ of the quadrotor (circle in bottom left axis).  c) For the $\uax$ output variable (white rectangle on the left of the frame): the ground truth value (green triangle); the output of A1, A2, A3 respectively (blue squares, from left to right).  d,e,f) A representation of output variables with the same notation: $\uay$ (bottom), $\uvz$ (right), $\uoz$ (top of the frame).
    }
    \label{fig:dataset}
\end{figure*}

\section{Experimental Setup}
\label{sec:setup}
\subsection{Hardware and Infrastructure}
All experiments take place in a
\SI[product-units = single]{10 x 10}{\metre} room fitted with an optical motion capture system.  Within the room, a flying area of \SI[product-units = single]{7 x 7}{\metre} is defined and fenced by a black lightweight net.

We use a Parrot Bebop 2 quadrotor; the quadrotor is networked via wifi to a server, and controlled through a ROS interface~\cite{bebop_autonomy}, which exposes at \SI{5}{\hertz} the velocity of the drone $\xodom$ computed from the visual odometry of a bottom-looking camera and, at \SI{30}{\hertz}, the front-facing camera feed. The drone uses a sonar and an IMU to estimate altitude and attitude; an onboard low-level controller accepts inputs $\uall$ and updates the  motors speed.
%At any time, before forwarding control commands to the drone, for safety reasons, we check that they won't make the drone exit the delimited area~\cite{drone_arena}.

The quadrotor is outfitted with a mocap target, so that its exact 6D pose is acquired in real time.  Moreover, we outfit the user's head with another mocap target; this is implemented by having the user wear one of two objects on their head, on top of which the mocap target is fixed: either a black baseball hat, or a thin elastic band.  While the former is clearly visible when the user is seen from the front, the latter is almost invisible. %(see Figure FIXME).  The setup allows us to acquire in real time the absolute 6D pose of both the quadrotor and the top of the user's head.

\subsection{Controller}
\label{sec:setup:controller}
We implement a simple baseline controller $C$ as a stateless function $f_{\controller}$. From the estimation of the person's pose, we compute a target point $\bar{p} = (s_x, s_y, s_z) +  \Delta e(\pi+s_\theta)$ in front them, at distance $\Delta$ along  unit vector $e(\pi+s_\theta)$. Then, we compute a desired velocity $\bar{v} =\left[ \frac{\bar p}{2 \tau} \right]^{v_{\max}}_{-v_{\max}}$ to reach $\bar p$ in $2 \tau$ time, limiting the components magnitude to $v_{\max}$. Finally, we compute a control output to reach velocity $\bar{v}$ and rotate towards the user in $\tau$ time:
\label{sec:controller}
\begin{align*}
    (\uax, \uax) &=\left[ \left(\left(\bar{v}_x, \bar{v}_y \right) - \left(v_x, v_y \right)\right)/\tau \right]^{a_{\max}}_{-a_{\max}}\\
    \uvz &= \bar{v}_z \\
    \uoz  &=\left[ \theta(\bar p) / \tau  \right]^{\omega_{\max}}_{-\omega_{\max}},
\end{align*}
where $\theta(\bar p)$ is the azimuth of vector $\bar p$.
Parameters are fixed to $ \tau = \SI{0.5}{\second},\, v_{\max} = \SI{1.5}{\metre \per \second},\, a_{\max} = \SI{1.0}{\metre \per \square \second},\, \omega_{\max} = \SI{2.0}{\radian \per \second}$.

%\theta(.) is the polar angle of the vector (in cylindrical coordinates)
% e(\theta) is the unit vector in direction \theta (in cylindrical coordinates)
% \tau = 0.5 s in our experiments 
% a_max = 1
% v_max = 1.5
% \omega_max = 2

\subsection{Dataset Acquisition}
\label{sec:setup:dataset}
In order to ease the acquisition of the dataset, the drone follows a controller $C_\text{acq}$ that, given the pose of the user and the pose of the quadrotor provided by the mocap system, generates control signals that keep the quadrotor at a parametrized distance (that we adjusted between 1.0 and 2.0 meters during the acquisitions) in front of the user's head. Such a controller is similar of the one described in Section~\ref{sec:controller}.%, that also accounts for the user's head velocity in order to anticipate their movements.
The outputs of $C_\text{acq}$ are not part of the dataset: they are only used to facilitate the acquisition of the sessions and make for an engaging experience for our test subjects.

% In order to ease the acquisition of the dataset, we implemented a simple controller $C_\text{acq}$ that, given the pose of the user and the pose of the quadrotor provided by the mocap system, generates control signals in order to keep the quadrotor at a fixed distance in front of the user's head.  Such a controller is similar to the one described in Section~\ref{sec:controller}, but also accounts for the user's head velocity in order to anticipate their movements. 

%; in fact, some sections of the recording were performed with the drone hovering in place or slowly rotating in place.

We recorded 15 different sessions, each with a different user (age 23 to 38, different ethnicity, height ranging from 160 to 197 cm, variable levels of physical fitness, different clothing styles, hairstyles and colors).  During the acquisition, users are instructed to move around the room freely: the quadrotor will then fly to stay in front of their head.  After an initial period with cautious motion, users start challenging the controller and moving rather aggressively (videos of recording sessions are available as supplementary material), so that the quadrotor struggles to keep up; this ensures that many different relative positions for the head and the quadrotor are represented in the data.  In some recordings, the users wear the hat, while in others the headband (which is almost not visible in the camera frames).  The room is equipped with 4 distinct light sources (2 overhead, 2 movable spotlights placed in different corners of the room), which we toggle and move at different times of the recording to add variability to the scene.  During some of the recordings, more people are in the flying area; moreover, often the quadrotor camera also sees people and different background objects standing outside of the flying area, such as computers, screens, desks, and windows.  While acquiring the data, we take care that, if more than one person is visible in the frame, the person wearing the mocap target is the one closest to the camera.  On occasion, the video link is temporarily corrupted and some frames are acquired with visual artifacts; we purposefully do not remove such frames from the recording as most of them can be still understood by an human observer. On average each session is $3$ minutes long, totaling more than $45$ minutes of flight.

All sessions are recorded in ROS bagfiles, from which we extract 79k dataset instances at a rate of 30Hz.  Each instance contains: the video frame $\xim$, the pose of the user head relative to the drone $\spose$, and the drone velocity $\xodom = \sodom = (v_x, v_y)$.  The controller $\controller$ is then applied to $(\spose,\sodom)$ to obtain the ground truth value $\uall$ (containing the four control variables).  Note that the ground truth control signal does not necessarily correspond to the control signal that the drone was receiving during the acquisition, which came from a different controller.

We use all data from three sessions (16k instances) as a test set to quantitatively evaluate the performance of the three approaches.  We randomly split the 63k instances from the remaining 12 sessions into a training set (50k instances) and a validation set (13k instances).

\subsection{Machine Learning Models}
\label{sec:setup:ML}
\subsubsection{Model M1}
Model M1 is implemented as a deep neural network using a ResNet~\cite{resnet} architecture, which has been recently adopted for similar quadrotor control tasks~\cite{Loquercio2018}.

The network accepts as input a $108 \times 60 \times 3$ RGB image, and produces the 4 components of $\sall$.  The internal architecture is a ResNet-8~\cite{resnet} followed by two dense layers with 256 and 128 neurons, respectively. %ResNet-8 is composed by a convolutional layer (64 @ $7\times7$ filters with stride $2$), a max-pooling layer ($3\times3$ with stride $2$) and 3 pre-activated ResNet blocks. Each pre-activated block is composed by two pre-activated convolutional layer with a number of $3\times3$ filter that increase for each block (i.e. the first block has 64 filters per layer, the second block 128 and the third 256), in addition the first layers of the second and third block have a stride of 2.

\subsubsection{Model M2}
Model M2 has a similar architecture as M1, but additionally accepts $\xodom$ through two additional input neurons that skip the ResNet-8 layers. These neurons, concatenated to the ResNet-8 output, are input to the two dense layers with 256 and 128 neurons.  The outputs correspond to the 4 components of $\uall$.

\subsubsection{Model M3}
Model M3 is implemented through a simple multilayer perceptron that maps 6 input values ($\sall^\text{A3} = \sall^\text{A1} = (\spose^\text{A1}, \sodom)$) to the 4 components of $\uall$; it contains 2 hidden layers, with 256 and 128 neurons respectively.
%This simple architecture is motivated by the fact that the function that this model has to approximate (i.e. the function implemented by $\controller$ )is not as complex as M1 or M2 tasks.

%\begin{figure*} % thpb
 %   \centering
 %   \includegraphics[width=0.51\textwidth]{img/architecture.png}%
 %   \caption{A representation of the three models used. M1(left), M2(center) and M3(right)}
 %   \label{fig:architectures}
%\end{figure*}

%\subsubsection{Data augmentation}
%\label{sec:setup:data_augmentation}
% In order to avoid biases in the datasets and make the models more robust, we use a data augmentation process before a datapoint is fed to one of the models. With a probability of 0.5 the datapoint is horizontally flipped.
%For model M1 it means that the image is horizontally flipped and the y and yaw are reversed, as a change of the sign means a change of the direction. For M2 the image is flipped and the label variable z\_ang and y\_lin are reversed. For M3 the input relative pose variables y and yaw are reversed and the target control variables z\_ang and y\_lin are reversed.

%Each instance used for training is transformed horizontally flipped with probability 0.5; the relevant outputs are correspondingly negated ($\sy$ and $\st$).  % For model M1 this means that $\xim$ is horizontally flipped and that we negate the ground truth values for $\sy$ and $\st$.  For model M2, $\xim$ is horizontally flipped and the ground truth values for $\uay$ and $\uoz$ are negated.  For model M3, we negate the input variables $\sy$ and $\st$ and the ground truth values for $\uay$ and $\uoz$.

\subsubsection{Training}
We train the three models with the same setup: 1) we use Mean Absolute Error as loss function and ADAM~\cite{adam} as optimizer with a learning rate of 0.001; 2) we speed up learning by reducing the learning rate when the validation loss plateaus for more than 5 epochs; 3) we use early stopping (10 epochs of patience on the validation loss, with a maximum of 200 epochs).  % At the end of training, we consider the model with the lowest validation loss.

\begin{figure*}[t]
    \centering
    \includegraphics[width=\textwidth]{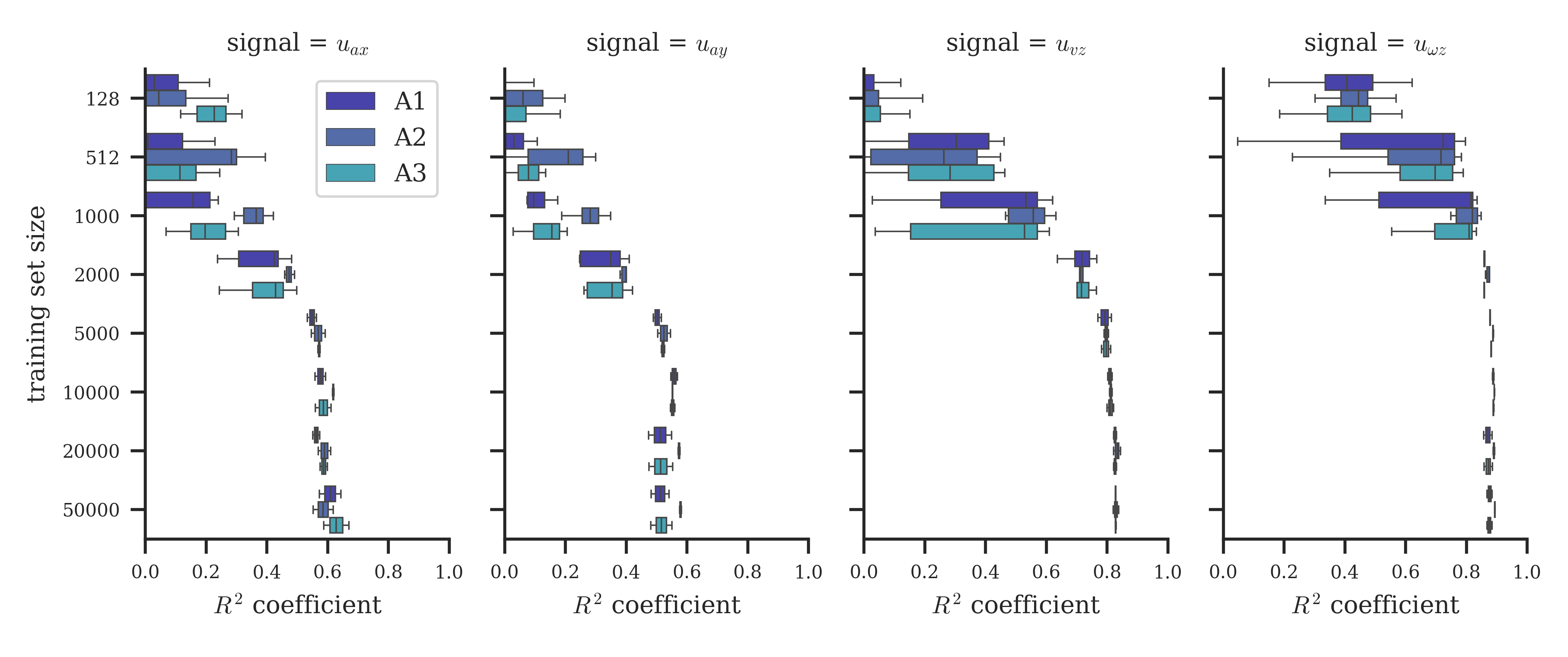}
    \caption{Quantitative results for the four control variables $\uall$: estimation quality versus number of training samples.}
    \label{fig:quantitative}
\end{figure*}

\begin{figure*}[t]
    \centering
    \includegraphics[width=\textwidth]{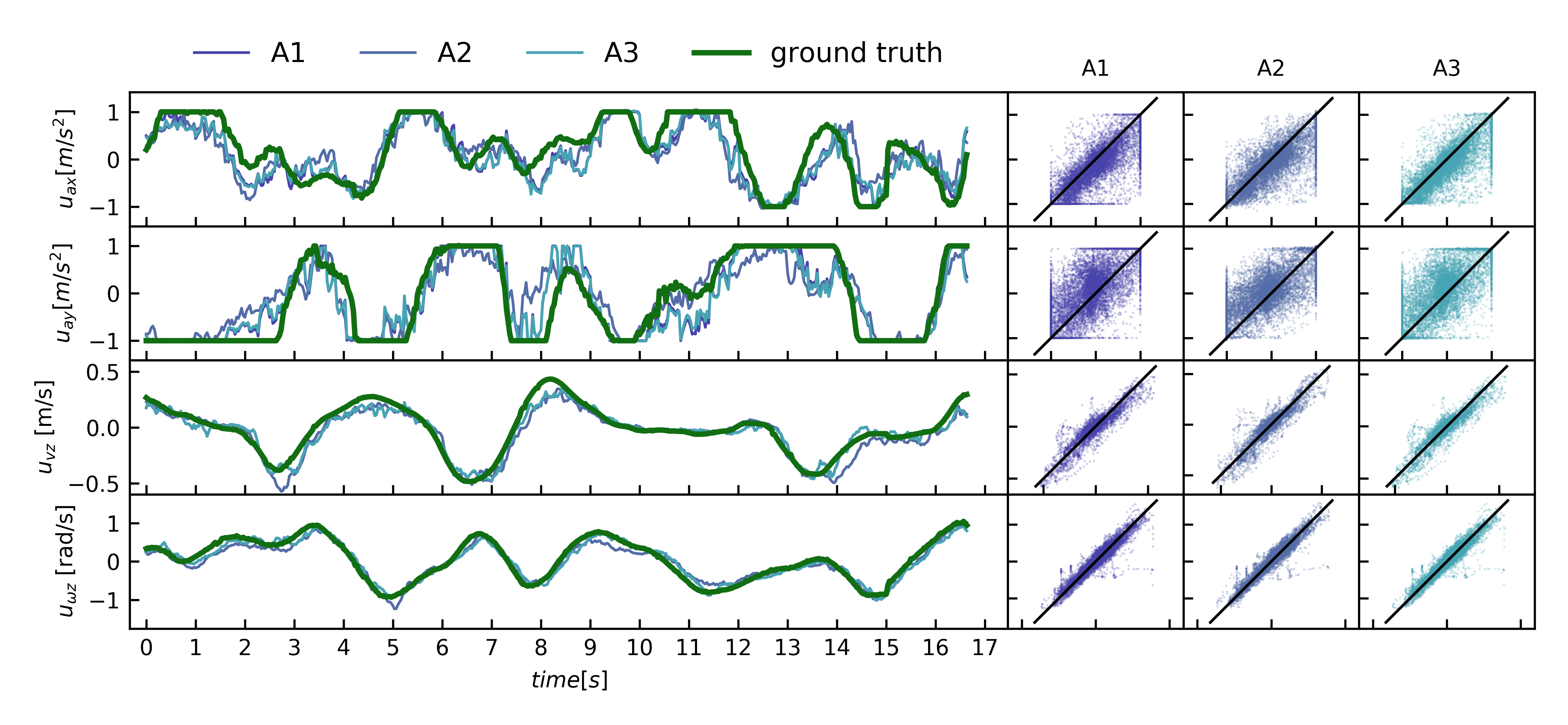}%
    \caption{We report each of the four control variables on a separate row. Left: the time series of the ground truth control (green thick line) and of the outputs of the ML approaches (thin lines), for a 16 second segment extracted from the test set, in which the user was moving very aggressively.  Right: the scatter plot comparing, on the y axis, the output of A1 (left), A2 (center), and A3 (right) to the ground truth (x axis), for the whole testing set; a perfect model would yield all points on the diagonal.}
    \label{fig:outputs}
\end{figure*}

\section{Experimental Results}
\label{sec:results}
We report three sets of experiments.  First, we quantitatively evaluate the prediction quality of different approaches for all instances in the testing set, versus the ground truth for $\uall$, also evaluating the impact of the training set size; second, we compare the trajectories of a quadrotor when it is controlled by an ideal controller, with the trajectories obtained by A1, A2 and A3.  Finally, we perform qualitative robustness tests on the system controlled by each of the three approaches.

\subsection{Quantitative results on testing instances}
For each component of $\uall^\text{A1}$, $\uall^\text{A2}$ and $\uall^\text{A3}$ we compute the coefficient of determination $R^2$~\cite{rsquared} of the estimate.
%The $R^2$ coefficient for a given output scalar variable $u \in \uall$ is computed as $$R^2 = \frac{\sum_i (u_i - \hat{u}_i)}{\sum_i (u_i - \bar{u})},$$ where $u_i$ denotes the ground truth value for the $i$th test instance, $\hat{u}_i$ its estimate, and $\bar{u}$ the average value of $u$.  The coefficient of determination 
This measure corresponds to the proportion of variance in $\uall$ that is explained by the model: a perfect estimator yields $R^2 = 1$; a dummy estimator that always returns the mean of the variable to be estimated yields $R^2 = 0$; even worse estimators yield $R^2 < 0$.  $R^2$ allows us to compare the quality of our estimate for different components of $\uall$ even though each has a different variance.

We compute these metrics for the three approaches trained on different amounts of training data; in particular, we are interested to compare how hard it is for the different approaches to achieve a given performance.  For a given training set size $T$, we randomly sample $T$ out of the 50k training instances, which we use to train the models for each of A1, A2 and A3.  Validation and testing datasets in all cases remain the same.  To account for the variability due the sampling of the training set (which is very large for low values of $T$), we repeat each experiment for up to 50 replicas. %$\{(128,50), (512,20), (1k, 10), (2k, 5), (5k, 2), (10k, 2), (20k, 2), (50k, 2)\}$.
\begin{comment}
\begin{table}\centering
\caption{Training size $T$ and number of replicas.}
\label{table:replicas}
\begin{tabular}{@{}lrrrrrrrr@{}}
\toprule
$T$        & 128  & 512  & 1k  & 2k & 5k & 10k & 20k & 50k \\ %\midrule
Replicas & 50 & 20 & 10 & 5 & 2 & 2 & 2 & 2 \\ \bottomrule
\end{tabular}
\end{table}
\end{comment}

Figure~\ref{fig:quantitative} reports the $R^2$ values resulting from the experiments described above.  We observe that prediction quality increases with $T$, but plateaus after $T=10000$; there is no clear difference in prediction performance among A1, A2 and A3.

$\uoz$ and $\uvz$ yield $R^2$ coefficients of $0.88$ and $0.82$, respectively. Prediction quality is significantly worse for $\uax$ and especially for $\uay$, with $R^2$ values of $0.59$ and $0.57$, respectively.  This is easily explained: $\uoz$ is mostly dependent on the position of the body in the frame, which is easy to perceive in $\xim$: $\uoz$ is learned significantly better than a dummy regressor, with as few as 128 training instances. 
%this sort of movement was also performed very frequently during dataset acquisition, so there is plenty of training data;  
$\uvz$ mostly depends on the vertical position of the head in the frame, which is also easy to perceive in $\xim$, but has less variability in the datasets.  Predicting $\uax$ and $\uay$ is harder: the former relies on an accurate perception of the distance of the user (which is confounded by their height and body size), and the latter on an estimate of the relative orientation of the head ($\st$), which is arguably hard to get on low resolution inputs.

Note that prediction quality is not necessarily related to the quality of the resulting robot behavior.  For example, a model which always yields a very tiny but systematic overestimate on $\uax$ yields an $R^2$ value close to 1, but would cause the quadrotor to crash on the user in a very short time.  On the contrary, a model that produces predictions affected by large amounts of uncorrelated noise with zero mean could yield acceptable behaviors (especially with short control timesteps) but very disappointing quantitative metrics.  To provide a better idea of the usability of the predictions for control, Figure~\ref{fig:outputs} illustrates the outputs of the three approaches (trained with $T=50000$) on a short segment extracted from a recording belonging to the testing set.
%, and scatter plots for each variable for each approach on the entire testing dataset.
We observe that the predicted signals closely track the ground truth; outputs for $\uax$ and $\uay$ exhibit a larger amount of high-frequency noise, which appears non-systematic.

\begin{figure}[t]
    \centering
    \vspace{1mm}
    \includegraphics[height=6cm]{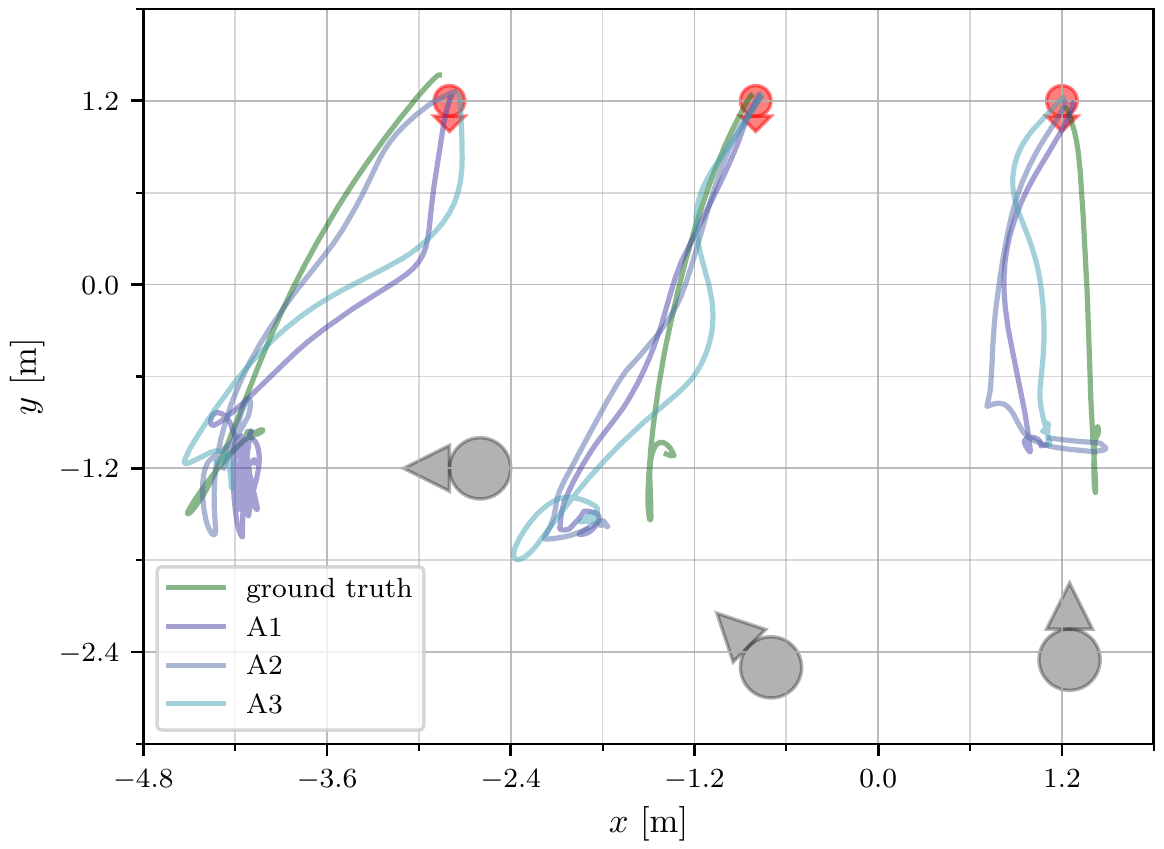}
    \caption{Trajectories followed by the quadrotor starting in the red pose while approaching a person standing still in different poses (black circles), from left to right: initially facing the quadrotor at \SI{90}{\degree}, \SI{45}{\degree}, and \SI{0}{\degree}. Trajectories represent one run and are colored by the controller in use.}
    \label{fig:traj_controllers}
\end{figure}

\begin{figure}[t]
    \centering
    \includegraphics[height=4.7cm]{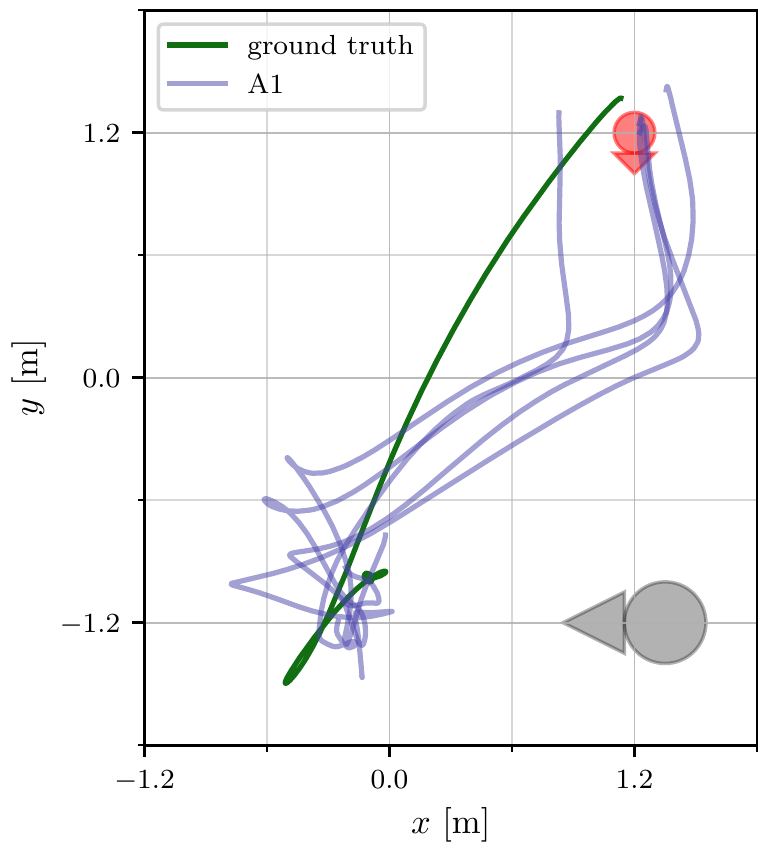}
    \includegraphics[height=4.7cm]{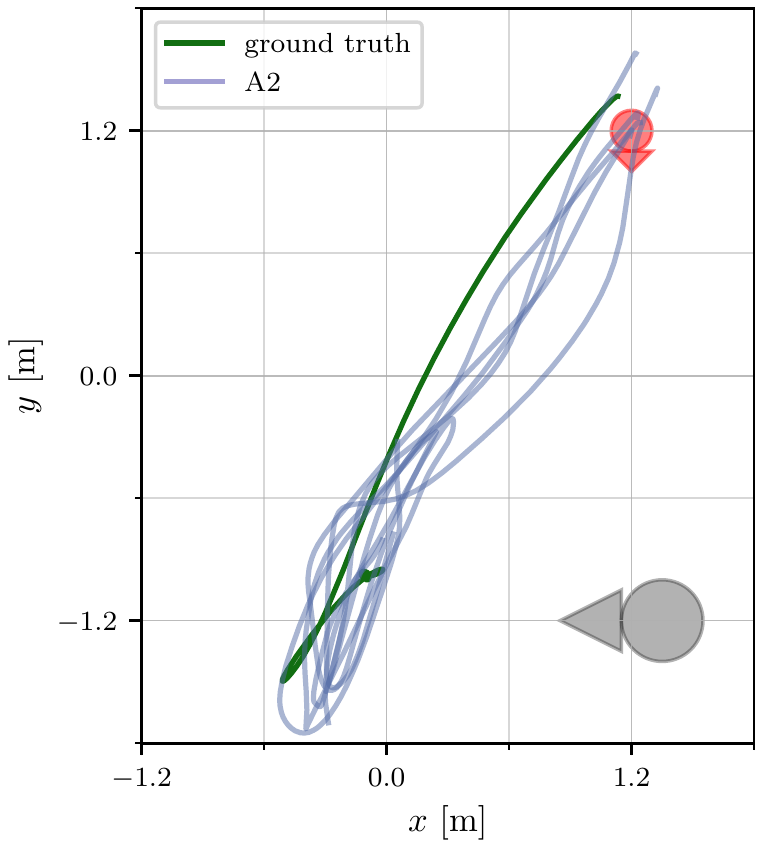}
    \caption{Five runs for each controller (left: A1, right: A2) towards a person initially facing the quadrotor at \SI{90}{\degree} and a single run with the ground truth controller (green).}
    \label{fig:traj}
\end{figure}

\subsection{Analysis of flying performance}

% \begin{figure*}[ht]
%     \centering
%     \begin{subfigure}[b]{0.45\textwidth}
%     \includegraphics[width=\textwidth]{img/girl1.pdf}
%     \caption{Trajectories towards three different poses of the person; from left to right: initially facing the quadrotor at \SI{90}{\degree}, \SI{45}{\degree}, and \SI{0}{\degree}.}
%     \label{fig:traj_controllers}
%     \end{subfigure}
%     \hfill
%     \begin{subfigure}[b]{0.45\textwidth}
%     \includegraphics[width=\textwidth]{img/girl2.pdf}
%     \caption{5 runs for each controller (left: A2, right: A1) towards a person initially facing the quadrotor at \SI{90}{\degree} and a single run with the ground truth controller (green).}
%     \label{fig:traj_repr}
%     \end{subfigure}
%     \caption{Trajectories followed by the quadrotor starting in the red pose while approaching a person standing still in different poses (black circles). Trajectories represent one run and are colored by the used controller.}
%     \label{fig:traj}
% \end{figure*}

Figure~\ref{fig:traj_controllers} compares the trajectories resulting from the ground truth controller, with those resulting from A1, A2, and A3. The quadrotor approaches a user (not in the dataset), who is standing still, from different relative poses: the final pose of the robot always faces the user, and is reached in about 5 seconds, after which the robot stabilizes in a short time.

In Figure~\ref{fig:traj}, the quadrotor's approach towards a different user (also not in the dataset) was run 5 times for A1 and A2.  This is the only case in our tests where we could notice a (small) difference in the behavior of different approaches, with A2 being somewhat smoother and closer to the ground truth trajectory.

% In Figure~\ref{fig:traj}
% each of A1 and A2 was run 5 times from approximately the same pose, with the user (not in the dataset) standing still: the final pose of the robot always faces the user, and is reached in about 5 seconds, after which the robot stabilizes in a short time. This is the only case in our tests where we could notice a (small) difference in the behavior of the three approaches, with A2 being somewhat smoother and closer to the ground truth trajectory.

% \begin{figure}[ht]
%     \centering
%     \includegraphics[width=\columnwidth]{img/girl2a.pdf}
%     \caption{Trajectories followed by the quadrotor starting in the red pose while approaching a person standing still and initially facing the quadrotor at \SI{90}{\degree} (black circles): 5 runs for each controller (left: A2, right: A1) and a single run with the ground truth controller (green).}
%     \label{fig:traj}
% \end{figure}

Supplementary videos report more details on this experiment and extensive tests where we challenge the system robustness by having multiple people in the frame, quick movements, distracting objects, sudden and extreme lighting variations.  During the tests, we cycled the control between the three approaches, and found that they behaved indistinguishably from each other; in all cases the drone behavior was predictable and perceived as safe by the users. % we show the system at work on users not belonging to the training or testing datasets.  In these experiments, we challenged the system in several ways, by having multiple people in the frame, quick movements, distracting objects, sudden and extreme lighting variations.  During the tests we cycled the control between the three approaches, and found that they behaved indistinguishably from each other; in all cases the drone behavior was predictable; in an episode, we turned off almost all lights of the room, which yielded almost black frames; in this case, the drone hovered in place.

%\begin{figure}[ht]
%    \centering
    %\includegraphics[width=0.24\textwidth]{img/screen_1.png}
    %\includegraphics[width=0.24\textwidth]{img/screen_2.png}
    %\includegraphics[width=0.24\textwidth]{img/screen_3.png}
    %\includegraphics[width=0.24\textwidth]{img/screen_4.png}
%    \includegraphics[width=0.15\textwidth]{img/screen_1.png}
%    \includegraphics[width=0.15\textwidth]{img/screen_2.png}
%    \includegraphics[width=0.15\textwidth]{img/screen_3.png}
    % \includegraphics[width=0.2\textwidth]{img/screen_4.png}
    % \begin{subfigure}[b]{0.3\textwidth}
    %     \includegraphics[width=\textwidth]{img/screen_1.png}
    %     \label{fig:screen_1}
    % \end{subfigure}
    % \begin{subfigure}[b]{0.3\textwidth}
    %     \includegraphics[width=\textwidth]{img/screen_2.png}
    %     \label{fig:screen_2}
    % \end{subfigure}\\[-2ex]
    % \begin{subfigure}[b]{0.3\textwidth}
    %     \includegraphics[width=\textwidth]{img/screen_3.png}
    %     \label{fig:screen_3}
    % \end{subfigure}
    %     \begin{subfigure}[b]{0.3\textwidth}
    %     \includegraphics[width=\textwidth]{img/screen_4.png}
    %     \label{fig:screen_4}
    % \end{subfigure}
%    \caption{Testing the robustness of the models (the video is available as supplementary material)}
%    \label{fig:trailer}
%\end{figure}

\section{Conclusions}
\label{sec:conclusions}
We considered the task of controlling a quadrotor to hover in front of a freely moving user using input data from an onboard camera.  On this task, we compared mediated approaches to end-to-end approaches; we found equivalent quantitative performance, learning difficulty, perceived quality of the robot behaviors and robustness to challenging inputs; in only one occasion we measured repeatably different trajectories, which were slightly smoother in the end-to-end approach, but correct in all cases and perceived as very similar.


\balance
\bibliographystyle{IEEEtran}
\bibliography{main} 

\end{document}